%% file: main.tex
\documentclass[numbered]{trbunofficial}
\usepackage{graphicx}
\usepackage{subcaption}
\usepackage{siunitx}
\usepackage{float}
\usepackage[table]{xcolor}
\usepackage[hidelinks]{hyperref}

\AuthorHeaders{Yao, Calvert, and Hoogendoorn}
\title{Driving pattern interpretation based on action phases clustering}

\author{%
  \textbf{Xue Yao, Corresponding Author}\\
  Department of Transport \& Planning, Delft University of Technology, Delft, the Netherlands\\
  Email: X.Yao-3@tudelft.nl\\
  \hfill\break
  \textbf{Simeon C. Calvert}\\
  Department of Transport \& Planning, Delft University of Technology, Delft, the Netherlands\\
  Email: s.c.calvert@tudelft.nl\\
  \hfill\break%
  \textbf{Serge P. Hoogendoorn}\\
  Department of Transport \& Planning, Delft University of Technology, Delft, the Netherlands\\
  Email: s.p.hoogendoorn@tudelft.nl\\
  \hfill\break%
}

\TotalWords{6048}

\begin{document}
\maketitle

\section{Abstract}
Current approaches to identifying driving heterogeneity face challenges in comprehending fundamental patterns from the perspective of underlying driving behavior mechanisms. The concept of \emph{Action phases} was proposed in our previous work, capturing the diversity of driving characteristics with physical meanings. This study presents a novel framework to further interpret \emph{driving patterns} by classifying \emph{Action phases} in an unsupervised manner. In this framework, a Resampling and Downsampling Method (RDM) is first applied to standardize the length of \emph{Action phases}. Then the clustering calibration procedure including ``Feature Selection'', ``Clustering Analysis'', ``Difference/Similarity Evaluation'', and ``\emph{Action phases} Re-extraction'' is iteratively applied until all differences among clusters and similarities within clusters reach the pre-determined criteria. Application of the framework using real-world datasets revealed six \emph{driving patterns} in the I80 dataset, labeled as ``Catch up'', ``Keep away'', and ``Maintain distance'', with both ``Stable'' and ``Unstable'' states. Notably, Unstable patterns are more numerous than Stable ones. ``Maintain distance'' is the most common among Stable patterns. These observations align with the dynamic nature of driving. Two patterns ``Stable keep away'' and ``Unstable catch up'' are missing in the US101 dataset, which is in line with our expectations as this dataset was previously shown to have less heterogeneity. This demonstrates the potential of \emph{driving patterns} in describing driving heterogeneity. The proposed framework promises advantages in addressing label scarcity in supervised learning and enhancing tasks such as driving behavior modeling and driving trajectory prediction. 

\hfill\break%
\noindent\textit{Keywords}: Driving behavior interpretation, Driving pattern, Action phase, Clustering analysis, Clustering calibration procedure
\newpage

\section{Introduction}
\input{sections/1_Introduction}

\section{Methodology}
\input{sections/2_Methodology}

\section{Evaluation of the clustering calibration process}
\input{sections/3_Clusteringcalibrationprocess}

\section{Driving Pattern Interpretation}
\input{sections/4_DrivingpatternInterpretation}

\section{Concluding Remarks}
\input{sections/5_Conclusion}

\section{Acknowledgements}
This work is supported by the Department of Transport \&Planning at Delft University of Technology and Data Analysis \& Traffic Simulation Lab (DiTTLab). 

\section{AUTHOR CONTRIBUTIONS}
The authors confirm their contribution to the paper as follows: study conception and design: Xue Yao, Simeon C. Calvert, Serge P. Hoogendoorn; data collection: Xue Yao; analysis and interpretation of results: Xue Yao, Simeon C. Calvert; draft manuscript preparation: Xue Yao, Serge P. Hoogendoorn, Simeon C. Calvert. All authors reviewed the results and approved the final version of the manuscript.

\newpage

\bibliographystyle{trb}
\bibliography{reference}
\end{document}

%% file: sections/1_Introduction.tex
Driving heterogeneity, recognized as the differences in driving behaviors exhibited by different driver/vehicle combinations under similar conditions, is widely acknowledged \cite{ossen2006interdriver}. Studies have shown that heterogeneity in driving behavior can lead to a rise in traffic accidents, congestion, and emissions \cite{sun2021modeling, kerner2004spatial}. Further, user acceptance of autonomous vehicles (AVs) has been found to hinge on accurately comprehending and emulating driving heterogeneity of human-driven vehicles (HDVs), such as human drivers' preferred driving styles \cite{tavakoli2022multimodal}. Thus, understanding driving heterogeneity plays a significant role in enhancing traffic operations and enabling manufacturers to design safe and efficient automated vehicles at various levels.

Existing studies have addressed driving heterogeneity by personalizing driving styles based on driving behavior data such as vehicle kinematics variables (e.g., velocity and headway) and vehicle dynamics variables (e.g., braking and throttle opening), which enable categorizing drivers into several groups \cite{zou2022multivariate}. For instance, \citep{wang2017driving} classified drivers into two categories (i.e., normal, and aggressive) based on velocity and throttle opening data. In another study, overtaking maneuvers were identified as low-medium-high risk levels based on speed and distance between vehicles \cite{figueira2020proposal}. Other studies have utilized car-following model parameters to distinguish driving styles \cite{sun2021modeling}. While this method captures drivers' static driving characteristics, it is not capable to describe the inherent traits of driving behavior. This is because driving behavior is a dynamic decision-making process \cite{zou2022multivariate}, drivers may exhibit heterogeneous driving styles in different traffic scenarios. Even under the same traffic scenario, the same driver's behaviors might vary at different time intervals.

Some studies found that driving heterogeneity can be derived by decomposing driving behavior into distinct driving patterns. As driving behavior displays certain characteristics during the transition of driving maneuvers \cite{terada2010multi}. As such, some researchers \cite{bender2015unsupervised, liu2014visualization} segmented driving behavior data into \emph{primitives} with unique characteristics. Typically, these primitives referred to distinct driving patterns, specifically known as primitive driving patterns. By doing so, the characteristics of driving behavior can be accurately captured by corresponding the traffic environment and driving maneuvers. Using supervised learning approaches, different patterns were extracted and assigned with semantic labels (e.g., rapidly closing in, falling behind) by learning sample features like vehicle operating data \cite{zou2022multivariate}. However, pre-labeling tasks are labor-intensive, limiting the implementation of supervised learning technologies in driving pattern recognition \cite{ackerman2017drive}. As a result, there is a growing interest in semantic analysis using unsupervised techniques. \citep{higgs2014segmentation} identified a specific set of state-action clusters and employed them to characterize potential driving patterns of passenger car and truck drivers. Employing a hierarchical Dirichlet process-Hidden semi-Markov Model (HDP-HSMM), \citep{wang2018driving} extracted 75 primitive driving patterns from time series driving data. This method allows for the identification of a wider variability in driving behavior by encompassing different driving characteristics. However, an excessive number of patterns, for example, 75, may limit the categorization's effectiveness due to reduced interpretability. This expansive classification has limitations in fully clarifying fundamental driving behaviors and understanding driving heterogeneity. As a result, continued research in this field is required to overcome these challenges.

In our previous research \cite{YaoAction2023}, the concept of \emph{Action phases} was introduced to capture driving characteristics with physical meanings, aiming to pave the way for identifying driving heterogeneity. It expanded the scope of the ``action point'' \cite{knoop2015relation} by incorporating additional variables to provide more comprehensive information about driving behavior. The \emph{action trend} space for each driving behavior variable is represented as $S = \{I, D, H, L\}$, which denotes `Increasing', `Decreasing', `Stable in a high value', and `Stable in a low value', respectively. One \emph{Action phase} carries a label name with each variable having a single \emph{action trend} in it. All the \emph{Action phases} obtained from a certain dataset constitute the \emph{Action phases} Library of the dataset, representing all driving behavior characteristics under this specific traffic flow condition. However, the dispersion of variables can complicate the uniform interpretation of driving behavior. Furthermore, when additional variables are considered, the quantity of label names escalates rapidly. Consequently, many \emph{Action phase}, despite having different label names, might exhibit minor differences in driving behavior. In this situation, consolidating \emph{Action phase} with similar characteristics into a smaller number of patterns can assist in interpreting driving behavior through the analysis of group-specific characteristics.

To bridge these research gaps, this study presents a novel framework to enhance the comprehension of driving behavior by classifying \emph{Action phases} into various \emph{driving patterns}. The unique contributions of this study include: (i) By using \emph{Action phases} with physical meanings, this unsupervised learning method holds dual advantages in eliminating pre-defined bias and ensuring behaviourally interpretable results. (ii) The clustering calibration process greatly assists in deriving \emph{driving patterns} with clear categorization and high internal similarity. Evaluation using real-world datasets demonstrates various \emph{driving patterns} with unique characteristics, accurately reflecting empirically observed driving behaviors. The findings also indicate the prospective advantages of using \emph{driving patterns} to illustrate the heterogeneity in driving behavior.

%% file: sections/2_Methodology.tex
In this section, we first give an overview of the framework for interpreting \emph{driving patterns}. Then the methods and techniques employed in each step are explained.

\subsection{Overall framework}
\begin{figure}[!ht]
  \centering
  \includegraphics[width=0.99\textwidth]{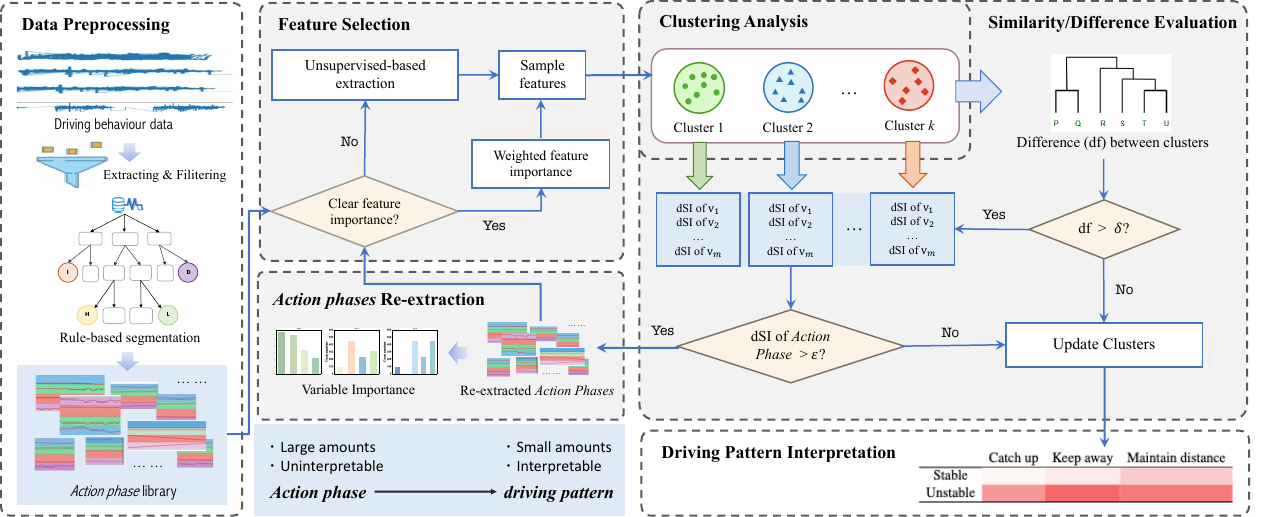}
  \caption{General framework of \emph{driving pattern} interpretation.}\label{fig:framework}
\end{figure}

The objective of the framework is to categorize \emph{Action phases} into several \emph{driving patterns} by utilizing an \emph{Action phase} Library. These patterns consist of different \emph{Action phases} that exhibit common traits. The \emph{Action phase} Library for each dataset was built in our previous work, which is briefly shown as the Data Preprocessing step in Figure~\ref{fig:framework}. Due to the limitations of the high-dimensional data used as input \cite{sun2021modeling}, Feature Selection is first conducted. If there is insufficient information regarding feature importance, particularly during the initial analysis phase, an unsupervised feature extraction method is utilized. After the variable importance is evaluated, the features are selected according to the weights of variable importance. 
The extracted features serve as input for the subsequent Cluster Analysis with an unknown cluster number $k$. The clustering results are evaluated from both inter-class and intra-class perspectives, demonstrated in the Similarity/Difference Evaluation step. In inter-class evaluation, the differences (df, indicating the distance) among clusters are calculated. As for intra-class, the shape similarities of variables between \emph{Action phases} within the same cluster are assessed using a dissimilarity index (dSI). Specifically, the inter-class evaluation takes precedence over the intra-class, implying that only clusters with substantial differences from each other are subjected to internal similarity evaluation. If the differences are less than a specific threshold $\delta$, the clustering results are directly accepted for \emph{driving pattern} description. In the intra-class evaluation, \emph{Action phases} that display a dSI greater than the threshold $\epsilon$ will be re-extracted and re-analysis in the next analysis round. The remaining \emph{Action phases}, which do not exceed this threshold, are maintained within the current cluster, contributing to the interpretation of \emph{driving patterns}. 

Notice that certain \emph{Action phases} are re-extracted based solely on one variable, while others are determined by multiple variables. This implies that some variables contribute significantly to the dissimilarity between \emph{Action phases} and have a more substantial role in the \emph{Action phases} Re-extraction process. We hypothesize that high-frequency variables carry more informative content reflecting driving behavior characteristics. These variables, therefore, take greater importance in explaining driving behavior. An importance score (IS) is subsequently computed to quantify the importance of driving behavior variables. Feature Selection, commencing from the second round of analysis, is guided by the variable IS obtained in the previous round, giving this process clear interpretive significance. 

The clustering calibration process of ``Feature Selection'', ``Clustering Analysis'', ``Difference/Similarity Evaluation'', and ``\emph{Action phases} Re-extraction'' (marked as gray background in Figure~\ref{fig:framework}) is iterated until all difference and similarity indices meet the pre-defined criteria. Ultimately, the optimal number of clusters is determined. Each cluster, representing a \emph{driving pattern}, is interpreted based on the characteristics of the \emph{Action phases} it contains. This framework enables the transformation of extensive sets of \emph{Action phases} into more concise \emph{driving patterns}, providing a semantic description of various driving behaviors. The techniques adopted at each stage will be detailed in the following subsections. 

\subsection{Data preprocessing}
The \emph{Action phases} accommodated driving trajectory segments with varying lengths in order to provide a more detailed representation of driving behavior characteristics. This refinement adds a layer of complexity to subsequent analyses, as most algorithms necessitate input data of equal length. Various methods have been suggested to mitigate this issue. One prevalent approach is \emph{padding}, where sequences are extended with a specific value (commonly zero) to match the length of the longest sequence. Alternatively, \emph{truncation} is utilized, wherein sequences are grouped into ``buckets'' of similar lengths (for instance, 1-10, 11-20, 21-30, etc.), followed by padding within each bucket. There are also studies that employed variable-length Recurrent Neural Networks (RNNs) to convert sequences into fixed-length representations.

Given that the time series data in \emph{Action phases} holds interpretable physical significance, the padded zeros also convey explicit meanings, such as $0 \mathrm{m/s}$ or $0 \mathrm{/s^{2}}$, which can significantly change the original characteristics of \emph{Action phases}. It is important to note that as the fundamental unit for describing driving behavior, the lengths of \emph{Action phases} can range from 2 seconds to 100 seconds, representing the duration of the driving behavior. Under this circumstance, the truncation method risks the loss of substantial information. The data processed by RNNs also suffers from a significant drawback - limited interpretability, which makes it unsuitable for data processing in this study.

\begin{figure}[!ht]
  \centering
  \includegraphics[width=0.65\textwidth]{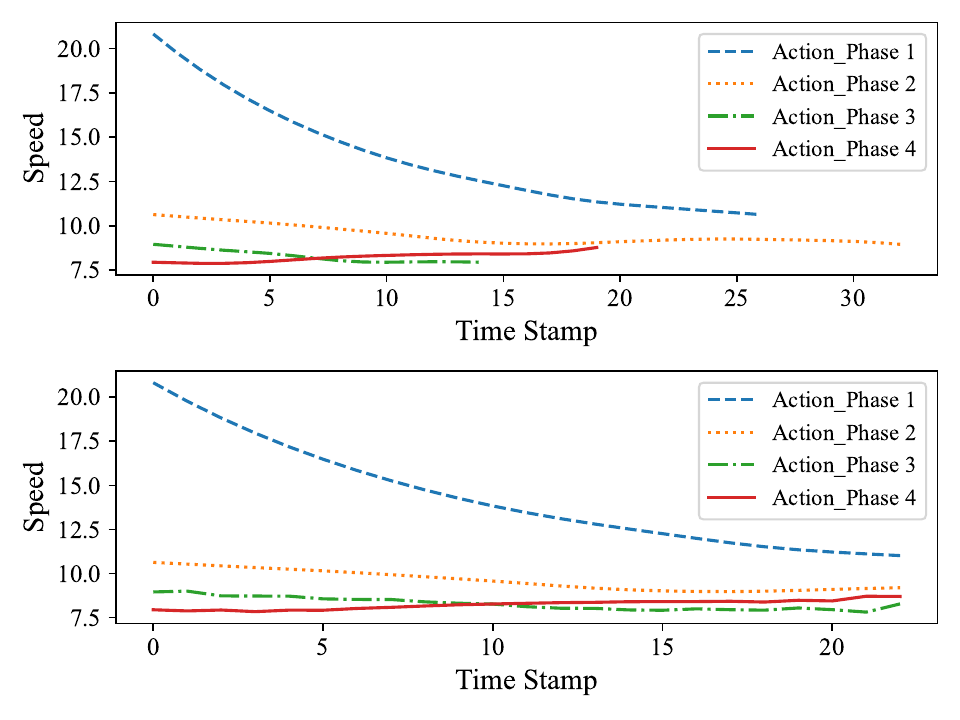}
  \caption{Fix \emph{Action Phase} length with Resampling and Downsampling Method (RDM).}\label{fig:resampledata}
\end{figure}

To standardize the length of input data while preserving the information of the original \emph{Action phases} to a large extent, the Resampling and Downsampling Method (RDM) is utilized. Initially, the median lengths of all \emph{Action phases} are determined to serve as a reference value. Then \emph{Action phases} shorter than this reference value are resampled to match the reference length using Fast Fourier Transform (FFT) and Inverse Fourier Transform (IFFT) \cite{702881}. In parallel, isometric extraction is implemented to truncate \emph{Action phases} exceeding the reference length down to the standard length. An example of fixing \emph{Action phase} length with RDM is illustrated in Figure~\ref{fig:resampledata}. 

\subsection{Feature Selection}
In the initial stage of clustering, an unsupervised feature extraction approach is employed due to the lack of clear information about the significance of variables. Principal Component Analysis (PCA) \cite{abdi2010principal}, Kernel PCA, t-SNE, among others, are frequently utilized methods for this purpose.

Upon completion of the first analysis round within the framework, the importance of variables can be evaluated. Consequently, for the subsequent stages of clustering, feature extraction is guided by the variable importance score (IS) acquired from the preceding analysis round. Borda Count \cite{kilgour2022weighted}, a highly recognized example of weighted scoring rules, is commonly applied in multi-candidate, single-winner electoral procedures. This method, which also provides the votes of each candidate during the selection of the winner, is adopted here for determining the importance of driving behavior variables.

Suppose the election has $m$ candidates (variables), $m = 1, 2, ..., M$, The ballots cast information is shown in matrix $X$, 

{\setlength\abovedisplayskip{0pt}
\setlength\belowdisplayskip{12pt}
\begin{linenomath}
  \begin{equation}
 X = 
 \begin{bmatrix}
x_{1,1}  &  x_{2,1}   & \cdots & x_{1,u}  \\
x_{2,1}  & x_{2,2}   & \cdots &  x_{2,u}  \\
\vdots & \ddots & \vdots \\
x_{m,1}  & x_{m,2}   & \cdots & x_{m,u}
\end{bmatrix}
  \end{equation}
\end{linenomath}}
here, $x_{m,u}$ signifies the count of the $u$-th ballots secured by the candidate $m$. Notably, ballot types are determined by the number of combined variables that re-extract the \emph{Action phases} demonstrating low similarity, hence $m=u$ in this situation.

The weight of each ballot type is represented by a \emph{score vector} $\alpha = (\alpha_1, \alpha_2, ..., \alpha_u)$ that satisfies $\alpha_1 \ge \alpha_2 \ge \ldots \alpha_u$. It is normalized and specifically set as $\{1, \frac{1}{2}, \frac{1}{3}, ..., \frac{1}{u}\}$ in the proposed framework.

Utilizing the ballot information from matrix $X$ and the score vector $\alpha$, the weighted Borda Score \cite{kilgour2022weighted} of variable $m \in V$ is computed by Equation~\ref{eq:bordaScore}. Then the importance score (IS) of variable $m \in V$ is obtained by normalizing the results.

{\setlength\abovedisplayskip{0pt}
\setlength\belowdisplayskip{12pt}
\begin{linenomath} 
  \begin{equation}\label{eq:bordaScore}
  Sc(m) = \sum_{i=1}^{u} \alpha x_i (m)
  \end{equation}
\end{linenomath}}

\subsection{Clustering Analysis}
The clustering analysis in our framework aims at finding typical \emph{driving patterns} in a given dataset. Various clustering approaches with inherent techniques have been proposed, this is due to the fact that there is no such precise definition to the notion of ``cluster'' \cite{rokach2005clustering}. According to \citep{fraley1998many}, the clustering approaches are divided into two different groups: hierarchical and partitioning techniques. Hierarchical clustering is chosen as the principal clustering method for two main reasons. The first is its capability to form a hierarchical depiction of the provided dataset, which in essence gives an outline of the distribution of \emph{driving patterns}. The second advantage is that hierarchical clustering offers reproducibility of the resulting clusters \cite{nguyen2019feature}. This mitigates the sensitivity to random initializations commonly encountered by partitioning clustering techniques such as k-means.

Hierarchical clustering can function in two ways - agglomerative (bottom-up) and divisive (top-down) - both of which are intrinsic strategies for building a binary tree. An agglomerative strategy is employed here, as it begins with each pattern as an individual cluster and inspects the connections between patterns or intermediate clusters. The fundamental concept is to merge the two closest patterns or intermediate clusters into a new, larger cluster. The proximity between any two patterns is determined using the Manhattan distance, also known as the city block distance \cite{stuart2016artificial}. When assessing the distance between two clusters, the average-link scheme is implemented, which measures the mean distance across all pairs of \emph{Action phases} from those clusters (refer to Equation~\ref{eq:hierarchicaltheory}). This procedure is repeated until only a single cluster is left, indicating all \emph{Action phases} have been grouped into the same cluster.

{\setlength\abovedisplayskip{0pt}
\setlength\belowdisplayskip{12pt}
\begin{linenomath} 
  \begin{equation}\label{eq:hierarchicaltheory}
  d(p, q) = \sum_{l=1}^{D} |p(l) - q(l) |
  \end{equation}
\end{linenomath}}

{\setlength\abovedisplayskip{0pt}
\setlength\belowdisplayskip{12pt}
\begin{linenomath}
  \begin{equation} \label{eq:distance}
  d(R, \mathcal{S}) = \frac{\sum_i^{N_{\mathcal{R}}} \sum_{j}^{N_{\mathcal{S}}} d \left(x_i^{\mathcal{R}}, x_j^{\mathcal{S}}\right)}{N_{\mathcal{R}} \times N_{\mathcal{S}}}
  \end{equation}
\end{linenomath}}
here, $p$ and $q$ denote two feature vectors that each represent a unique \emph{Action phase}, with $D$ as the total number of dimensions. $p(l)$ denotes the $l$-th element present within vector $p$. $\mathcal{R}$ and $\mathcal{S}$ symbolize two distinct clusters. $N_\mathcal{R}$ and $N_\mathcal{S}$ are indicative of the quantity of \emph{Action phases} in clusters $\mathcal{R}$ and $\mathcal{S}$ respectively. $x_l^\mathcal{R}$ stands for the feature vector that represents the $i$-th \emph{Action phase} located within cluster $\mathcal{R}$. $d(p, q)$ calculates the total sum of the absolute differences between corresponding elements of vectors $p$ and $q$, which is the standard computation for Manhattan distance.

The term $d(R, \mathcal{S})$ in Equation~\ref{eq:distance} computes the average distance between all possible pair combinations of \emph{Action phases} from clusters $\mathcal{R}$ and $\mathcal{S}$. This is achieved by adding up the distances between every possible pair and then dividing by the total number of such pairs. The total number of pairs is given by multiplying the count of \emph{Action phases} in each of the two clusters.

\subsection{DTW-based Similarity Evaluation}
The key to effective clustering is ensuring significant distances between clusters while also maintaining high internal similarity. Each variable in the \emph{Action phases} can be considered as a set of time-series data. However, the challenge with comparing the similarity of two time-series data points is that even if they share similar characteristics or trends, they might not align along the time axis. For instance, two sets of velocities could demonstrate similar trends over time, but one may occur at a faster or slower rate than the other. This is where Dynamic Time Warping (DTW) proves useful.

The main idea behind DTW is to compare the distances of two sequences under all possible ``warpings'', and then identify the optimal match among these warpings \cite{salvador2007toward}. Consider two sequences, $X=[x_1, x_2, ..., x_n]$ and $Y=[y_1, y_2, ..., y_m]$ of lengths $|X|$ and $|Y|$ respectively. A warp path $W$ is then created, as shown in Equation~\ref{eq:warpPath}.

{\setlength\abovedisplayskip{0pt}
\setlength\belowdisplayskip{12pt}
\begin{linenomath}
  \begin{equation} \label{eq:warpPath}
    W = w_1, w_2, ..., w_K \; \; \max(|X|,|Y|\leq K < |X|+|Y|)
  \end{equation}
\end{linenomath}}
here, $K$ signifies the length of the warp path, and the $k$-th element of the warp path is $w_k = (i,j)$, where $i$ and $j$ are indices of time series $X$ and $Y$, respectively. The warp path initiates at the beginning of each time series at $w_1 = (1, 1)$ and finishes at the end of both time series at $w_K = (|X|, |Y|)$. A constraint on the warp path mandates $i$ and $j$ to be monotonically increasing in the warp path. Every index of both time series must be engaged. This constraint can be expressed more formally as follows:

{\setlength\abovedisplayskip{0pt}
\setlength\belowdisplayskip{12pt}
\begin{linenomath}
  \begin{equation}
    w_k = (i, j), \ w_{k+1} = (i', j') \;  \; i \leq i' \leq i+1, j \leq j' \leq j+1
  \end{equation}
\end{linenomath}}

The optimal warp path is the one with minimum distance, where the distance (or cost) of a warp path $W$ is

{\setlength\abovedisplayskip{0pt}
\setlength\belowdisplayskip{12pt}
\begin{linenomath}
  \begin{equation}
    Dist(W) = \sum_{k=1}^{k=K} Dist(w_{ki}, w_{kj})
  \end{equation}
\end{linenomath}}
$Dist(w_{ki}, w_{kj})$ represents the distance between the two data point indices (one from $X$ and one from $Y$) in the $k$-th element of the warp path.

Dynamic programming is then deployed to identify this minimum-distance warp path between the two time series, providing the best match between them. This can be expressed as:

{\setlength\abovedisplayskip{0pt}
\setlength\belowdisplayskip{12pt}
\begin{linenomath}
  \begin{equation}
  D(i, j) = Dist(i, j) + \min[D(i-1, j), D(i, j-1), D(i-1,j-1)]
  \end{equation}
\end{linenomath}}

However, the complexity of this algorithm is $O(N^2)$. When the time series is considerably long, the efficiency of the DTW reduces, thus being unable to meet the needs. Consequently, FastDTW was developed, providing a linear and accurate approximation of dynamic time warping. FastDTW utilizes a multilevel strategy that recursively projects a warped path to a higher resolution and fine-tunes it. The three critical operations in this process include \cite{salvador2007toward}:

1) \emph{Coarsening} – Reducing a time series into a smaller time series that accurately represents the same curve with fewer data points.

2) \emph{Projection} – Identifying a minimum-distance warp path at a lower resolution, and utilizing it as an initial guess for a higher resolution's minimum-distance warp path.

3) \emph{Refinement} – Refining the warp path projected from a lower resolution by locally adjusting the warp path.

These operations effectively lower the time complexity to $O(N)$. While the strategy to reduce the search space may increase the error, these errors usually remain within an acceptable range \cite{salvador2007toward}.

%% file: sections/3_Clusteringcalibrationprocess.tex
Evaluation of the clustering calibration process is designed and presented in this section. First, data preparation including \emph{Action phases} review and feature selection results is introduced. Then, hierarchical clustering is conducted, and the results are evaluated by using a qualitative measure for inter-class difference and a quantitative measure for intra-class similarity. \emph{Action phases} demonstrating low intra-class similarities are extracted, during which process the variables' importance is calculated. 

\subsection{Data and Feature Selection}
As proposed in \cite{YaoAction2023}, driving behavior trajectories are segmented to yield \emph{Action phases}, with each driving variable in these phases displaying a single trend. For example, $(D, L, L, I)$ indicates the vehicle's velocity ($v$) has a trend of decreasing, the acceleration ($a$) and distance ($d$) are keeping a low value, and the speed difference ($\Delta v$) is increasing. All the \emph{Action phases} extracted from one dataset (representing a specific traffic flow condition) constitute the \emph{Action phase} Library of this dataset, which is adopted as the initial data in this study. The total size of the \emph{Action phase} Library amounts to 1764 for the I80 dataset and 13564 for the US101 dataset.

\begin{figure}[!ht]
  \centering
  \begin{subfigure}{0.45\textwidth}
  \centering
    \includegraphics[width=\textwidth]{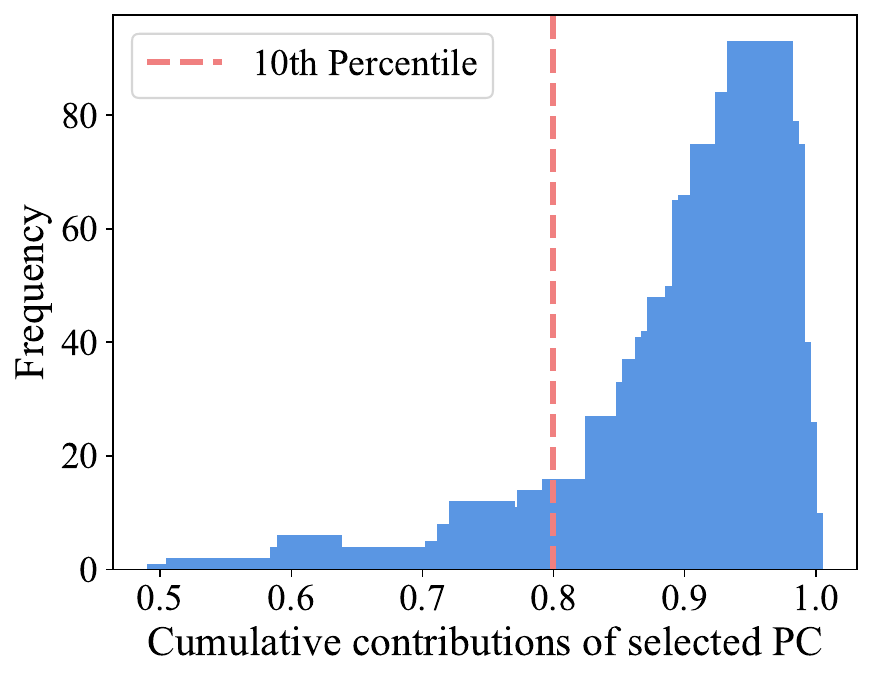}
    \caption{I80 dataset.}\label{fig:pca80}
  \end{subfigure}
  \hspace{0.1cm}
  \begin{subfigure}{0.45\textwidth}
  \centering
    \includegraphics[width=\textwidth]{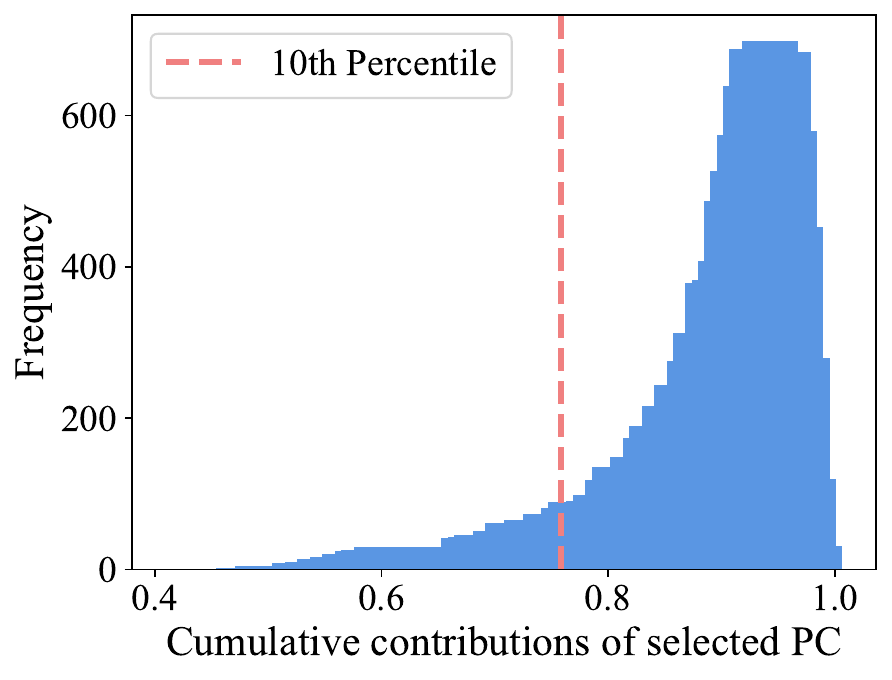}
    \caption{US101 dataset.}\label{fig:pca101}
  \end{subfigure}
  \caption{Distribution of PC1's cumulative contributions}\label{fig:PCA-1st}
\end{figure}

\emph{Action phases} consist of trajectory information using four variables, making feature extraction crucial for high-quality cluster analysis. This study employs Principal Component Analysis (PCA), an unsupervised feature selection method, to cohere variables and extract significant features. The results are displayed in Figure~\ref{fig:PCA-1st}. Notice that in the I80 dataset, the cumulative contribution of the first Principal Component (PC1) exceeds 80\% for 90\% of \emph{Action phases}, as indicated by the red dotted line in Figure~\ref{fig:pca80}. Similarly, PC1 has a considerable cumulative contribution of over 75.82\% in the US101 dataset, as shown in Figure~\ref{fig:pca101}. Consequently, PC1s are selected and used as the input for subsequent clustering analysis.

\subsection{Clustering Analysis}
Determining a suitable validation strategy for an unsupervised learning problem is acknowledged to be a complex task. The literature generally offers two validation criteria: internal and external \cite{rendon2011internal}. The external criterion compares the clustering outcome with existing knowledge of the dataset's structure (commonly referred to as true labels), whereas such information is usually subjective or unavailable. As a result, this study concentrates on the internal criterion, which assesses clustering results based on the inherent properties of the dataset. The evaluation undertaken in this study includes i) a qualitative measure for inter-class by visually observing the dendrogram to ascertain the differences between clusters, and ii) a quantitative measure for intra-class by calculating the similarity of \emph{Action phases} within a cluster.

\begin{figure}[!ht]
  \centering
  \begin{subfigure}{0.47\textwidth}
  \centering
    \includegraphics[width=\textwidth]{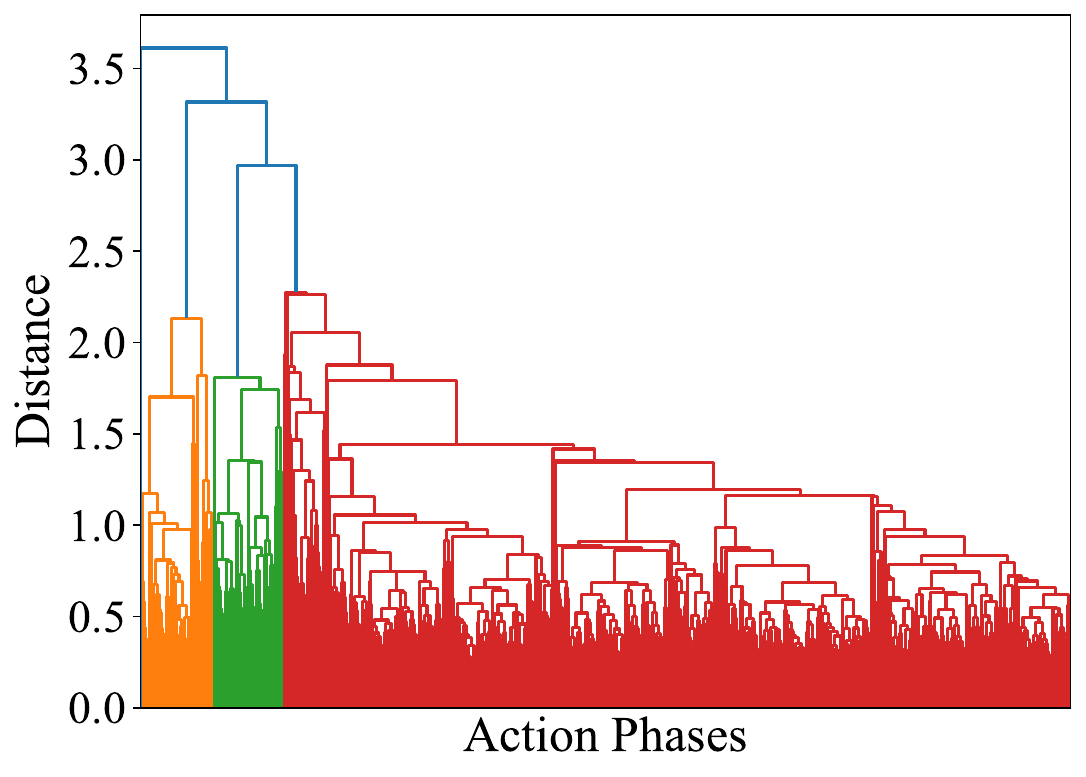}
    \caption{I80 dataset.}\label{fig:1stClustering-I80}
  \end{subfigure}
  \hspace{0.2cm}
  \begin{subfigure}{0.47\textwidth}
  \centering
    \includegraphics[width=\textwidth]{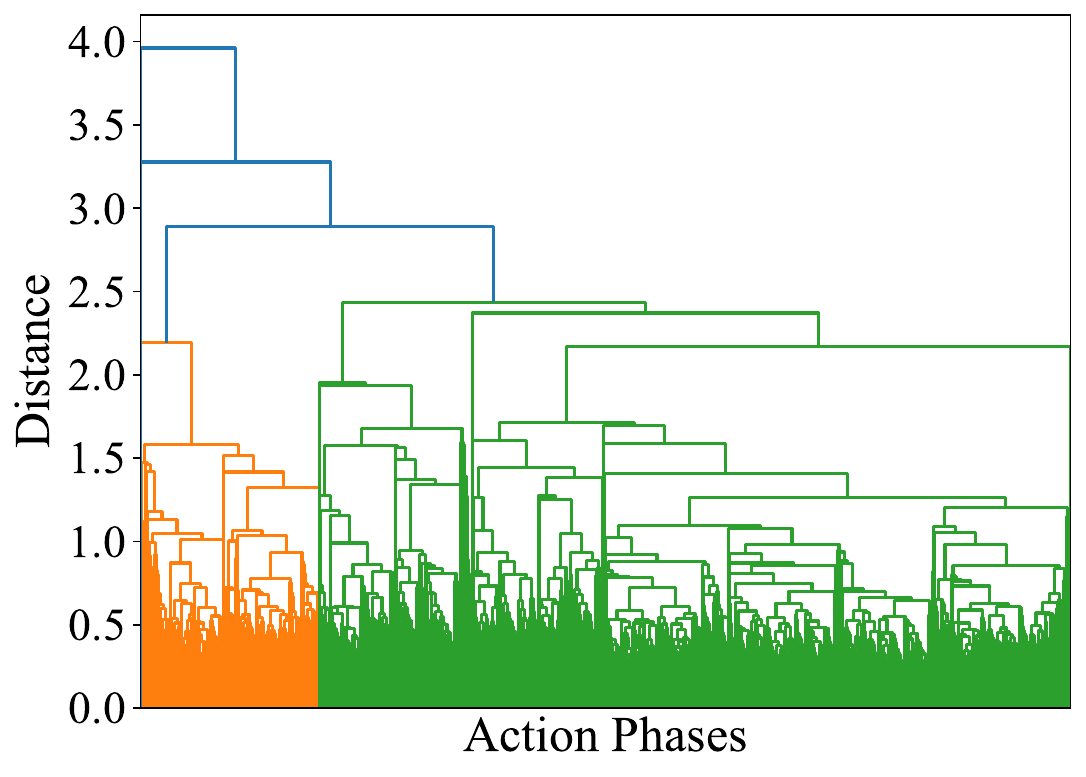}
    \caption{US101 dataset.}\label{fig:1stClustering-US101}
  \end{subfigure}
  \caption{Dendrogram representations of hierarchical clustering results.}\label{fig:1stClustering}
\end{figure}

\subsubsection{Qualitative evaluation}
Clustering results are evaluated by examining different (intermediate) clusters, i.e., branches, demonstrated by the two dendrograms shown in Figure~\ref{fig:1stClustering}. The metric is based on the similarities among the \emph{Action phases} within the same cluster. Focusing on the dendrogram for the I80 dataset depicted in Figure~\ref{fig:1stClustering-I80}, the first analysis involves the two highest sub-trees (or branches): The left sub-tree comprises small, simple patterns of \emph{Action phases}, as colored by orange; conversely, the right sub-tree displays a variety of patterns. Thus, further examination is conducted on the right sub-tree, revealing that (i) the left sub-subtree contains small, straightforward patterns of \emph{Action phases}; (ii) despite the right sub-subtree demonstrates a diversity of patterns, they occur at close distances. Following this analysis, three clusters are identified in the I80 dataset (marked as Cluster 3, 2, and 1), and two clusters are observed in the US101 dataset (marked as Cluster 2 and 1). As shown in Figure~\ref{fig:1stClustering-I80} and Figure~\ref{fig:1stClustering-US101}, the distance between clusters obtained from the two datasets exceeds 2, surpassing the pre-established threshold $\delta = 1$. Hence, the analysis continues with evaluating the similarity within the cluster. 

\subsubsection{Quantitative Evaluation}
\begin{figure}[!ht]
  \centering
  \includegraphics[width=0.97\textwidth]{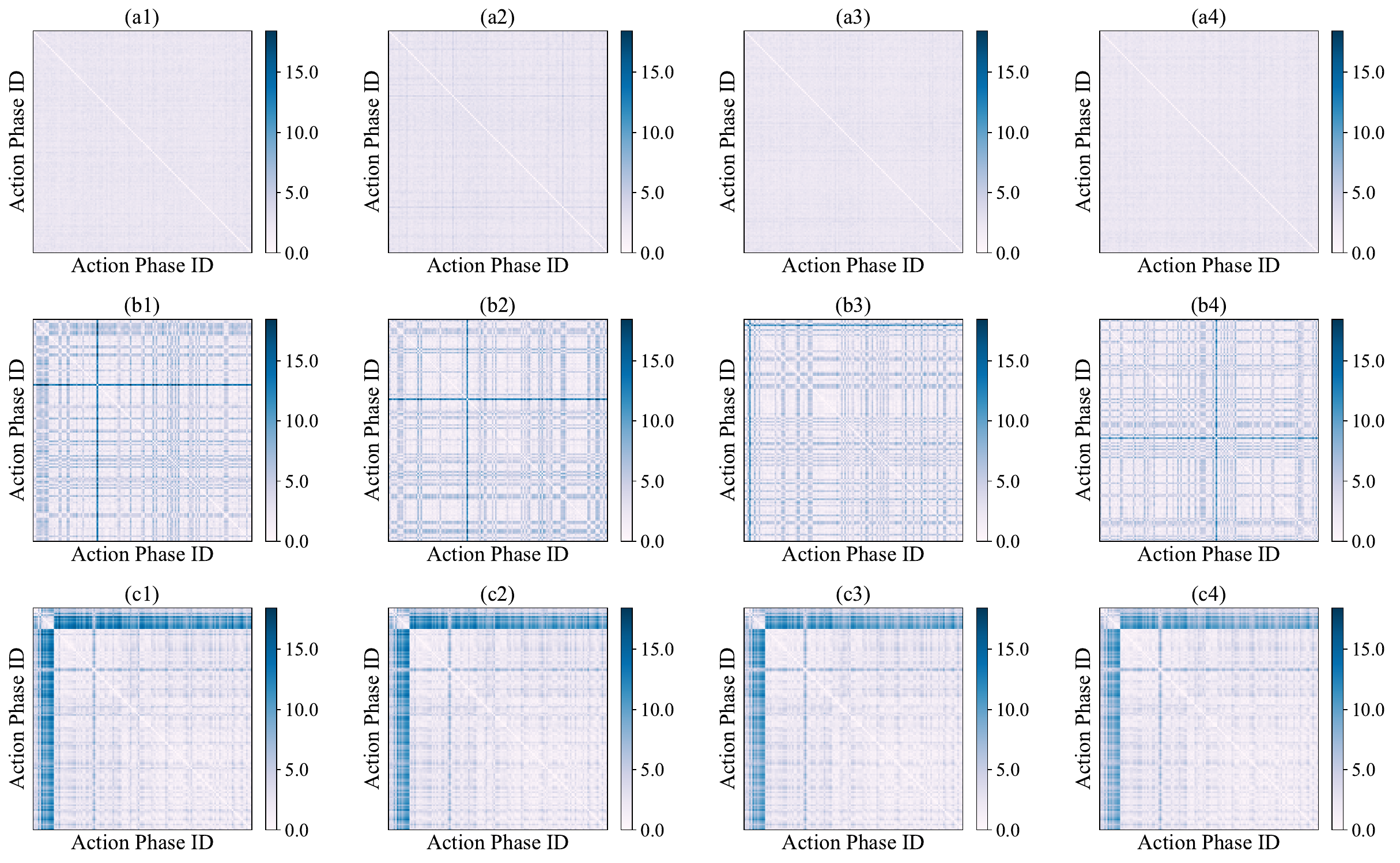}
  \caption{Dissimilarity Index according to fastDTW - I80 dataset.}\label{fig:DTW-similar-I80}
\end{figure}

\begin{figure}[!ht]
  \centering
  \includegraphics[width=0.97\textwidth]{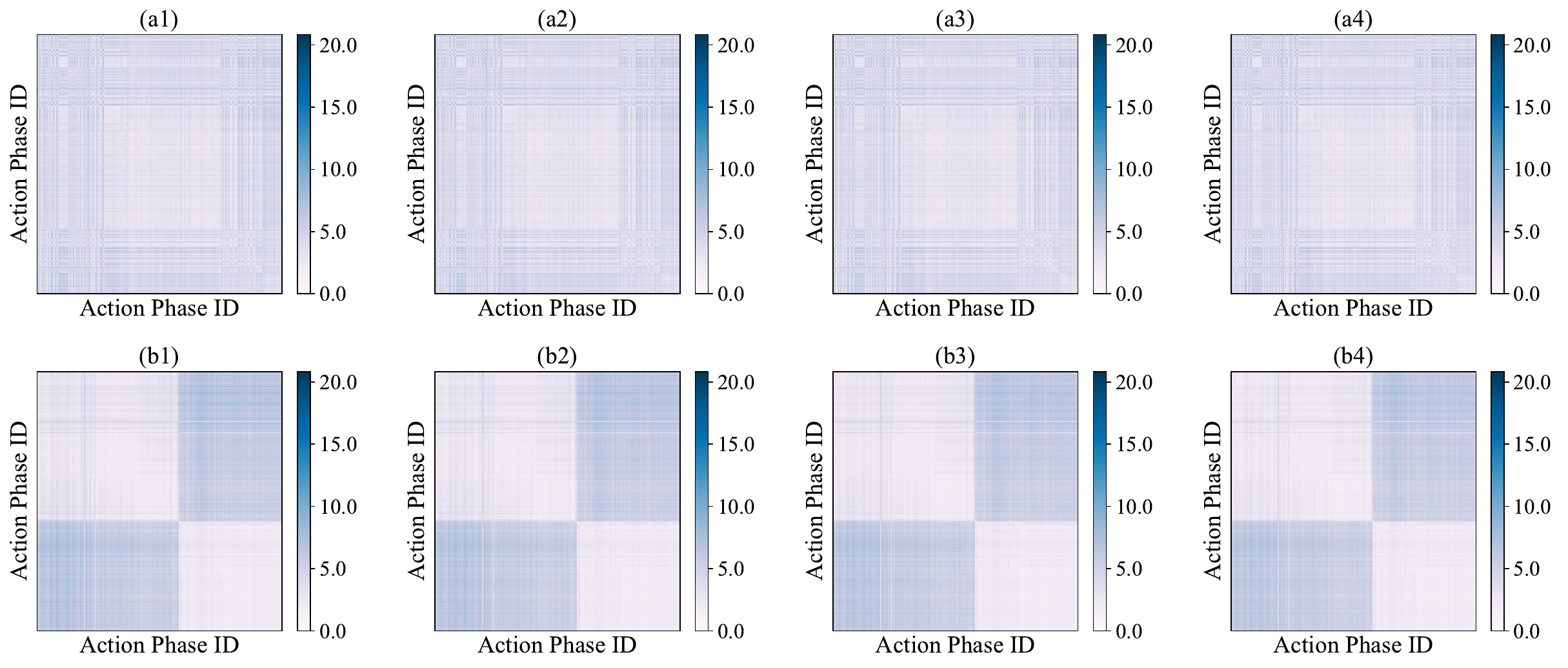}
  \caption{Dissimilarity Index according to fastDTW - US101 dataset.}\label{fig:DTW-similar-US101}
\end{figure}

FastDTW compares the shape similarity by calculating the minimum-distance warp path between two variable data sets. This minimum distance is used as a dissimilarity index (dSI) to provide a quantitative assessment of the clustering results. A value closer to 0 implies higher similarity between the two variable sequences. Conversely, a larger dSI suggests less similarity in their shape. The similarity assessment of the I80 dataset is depicted in Figure~\ref{fig:DTW-similar-I80}. (a) - (c) represent the three clusters obtained from hierarchical clustering, while 1 - 4 denotes the four variables $v$, $a$, $d$, and $\Delta v$ considered in this study. All figures use a uniform color bar as a reference, with darker colors indicating greater distances or less similarity. As represented by the deeper blue lines, larger dSI values are observed in both Cluster 2 and Cluster 3, suggesting these \emph{Action phases} have a relatively low similarity. In contrast, Cluster 1 generally has low dSI values, signifying a high degree of similarity between \emph{Action phases} in this cluster. The same analysis is also carried out on the US101 dataset, as shown in Figure~\ref{fig:DTW-similar-US101}.

To improve the intra-class similarity, \emph{Action phases} with a dSI index for variables exceeding a threshold value $\epsilon$ will be re-extracted for the next round of analysis. In this context, the threshold $\epsilon$ is defined as the 99th percentile of all dSI values in a given dataset. Consequently, a total of 1054 and 11342 \emph{Action phases} are re-extracted from the I80 and US101 datasets, respectively. More specifically, from the I80 dataset, 1089 \emph{Action phases} are extracted from Cluster 1, while 126 and 133 \emph{Action Phases} are from Cluster 2 and Cluster 3 respectively. As for the US101 dataset, the numbers are 9258 and 2084 for Cluster 1 and Cluster 2, respectively.

\subsection{Variable Importance Evaluation}
\begin{figure}[!ht]
  \centering
  \includegraphics[width=0.95\textwidth]{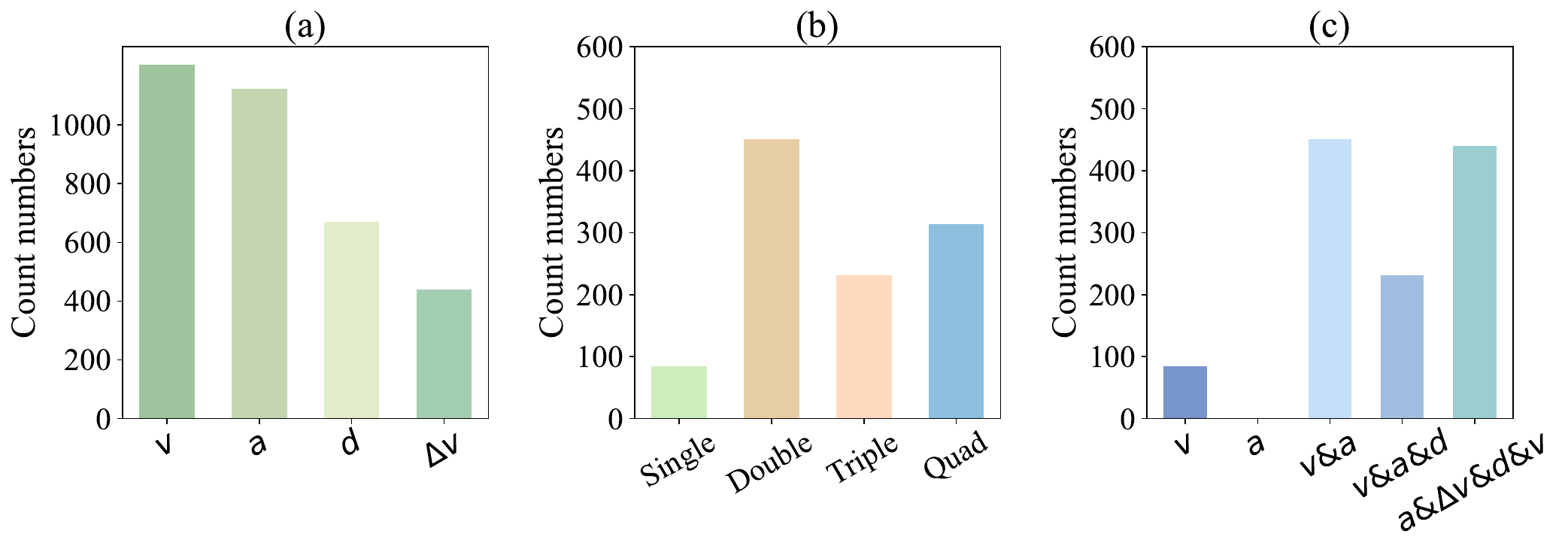}
  \caption{Statistics of variable contribution in re-extracting \emph{Action Phases} - I80 dataset.}\label{fig:variable-importance-I80}
\end{figure}

\begin{figure}[!ht]
  \centering
  \includegraphics[width=0.95\textwidth]{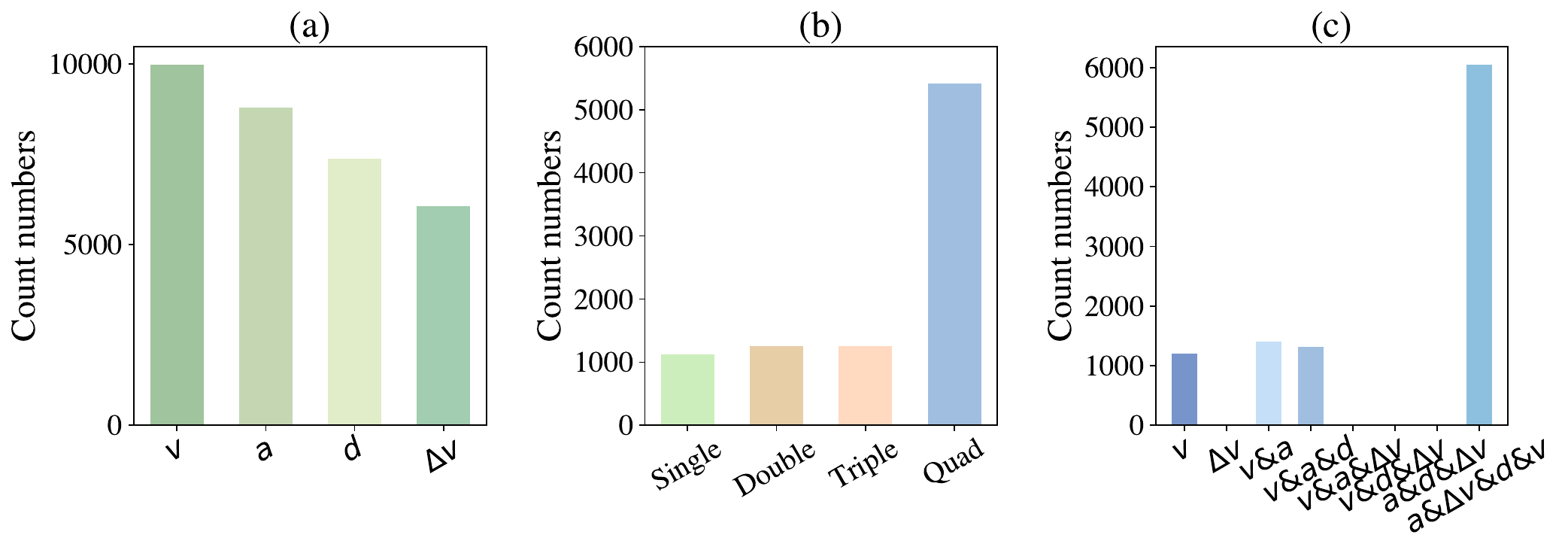}
  \caption{Statistics of variable contribution in re-extracting \emph{Action Phases} - US101 dataset.}\label{fig:variable-importance-US101}
\end{figure}

Figure~\ref{fig:variable-importance-I80}(a) and Figure~\ref{fig:variable-importance-US101}(a) display the count of each variable involved in the re-extraction process. The variable $v$ appears to be the most recurrent one used in \emph{Action phases} re-extraction in both datasets, succeeded by $a$, $d$, and $\Delta v$, in that sequence. It is worth noting that an \emph{Action phase} could be selected due to the marked dissimilarity in one or multiple variables, as the statistical data shown in Figure~\ref{fig:variable-importance-I80}(b) and Figure~\ref{fig:variable-importance-US101}(b). In the I80 dataset, the selection of most \emph{Action phases} relies on two variables, whereas in the US101 dataset, all four variables commonly participate. Figure~\ref{fig:variable-importance-I80}(c) and Figure~\ref{fig:variable-importance-US101}(c) illustrate various combinations of these variables used in the re-extraction process. The combination of all four variables displays high frequency in both datasets. When it comes to univariate extraction, velocity $v$ exhibits the highest frequency, and for bivariate extraction, the combination of $v$ and $a$ prevails.

In this study, the $m$ candidates in Borda count are the four variables, namely $v$, $a$, $d$, and $\Delta v$. It is assumed that a variable that is individually involved more in the re-extraction process holds greater significance. As such, four different types of ballots are generated, leading to the \emph{score vector} $\alpha = \{1, \frac{1}{2}, \frac{1}{3}, \frac{1}{4}\}$. The ballots matrix $X$ for these variables is displayed in Table~\ref{tab:VIscore-I80} and Table~\ref{tab:VIscore-US101}. Subsequently, the weighted Borda Score (wBS) for each variable is calculated and normalized to yield the Importance Score (IS) of driving behavior variables. The outcomes are $IS_{80} = [1.0, 0.833, 0.831, 0.223]$ and $IS_{101} = [1.0, 0.674, 0.532, 0.411]$. Higher scores represent higher importance. This result will serve as a guide for selecting features of \emph{Action phases} in the subsequent round of analysis.

\begin{table}[!ht]
	\caption{Variable ballot count - I80 dataset}\label{tab:VIscore-I80}
	\begin{center}
		\begin{tabular}{l l l l l}
		Combination Num. & Velocity($v)$ & Acceleration($a$) & Distance($d$) & Speed difference($\Delta v$)\\\hline
		1 &	84   & 1 & 0 & 0  \\
		2 &	451  & 451 & 0 & 0  \\
		3 &	231  & 231 & 231 & 0 \\
		4 &	440  & 440 & 440  & 440 \\\hline
		\end{tabular}
	\end{center}
\end{table}

\begin{table}[!ht]
	\caption{Variable ballot count - US101 dataset}\label{tab:VIscore-US101}
	\begin{center}
		\begin{tabular}{l l l l l}
		Combination Num. & Velocity($v)$ & Acceleration($a$) & Distance($d$) & Speed difference($\Delta v$)\\\hline
		1 &	1199   & 0 & 1 & 0  \\
		2 &	1046  & 1046 & 0 & 0  \\
		3 &	1325  & 1328 & 1325 & 0 \\
		4 &	6050  & 6050 & 6050  & 6050 \\\hline
		\end{tabular}
	\end{center}
\end{table}

\subsection{Analysis on Re-extracted \emph{Action phases}}
\begin{figure}[!ht]
  \centering
  \begin{subfigure}{0.46\textwidth}
  \centering
    \includegraphics[width=\textwidth]{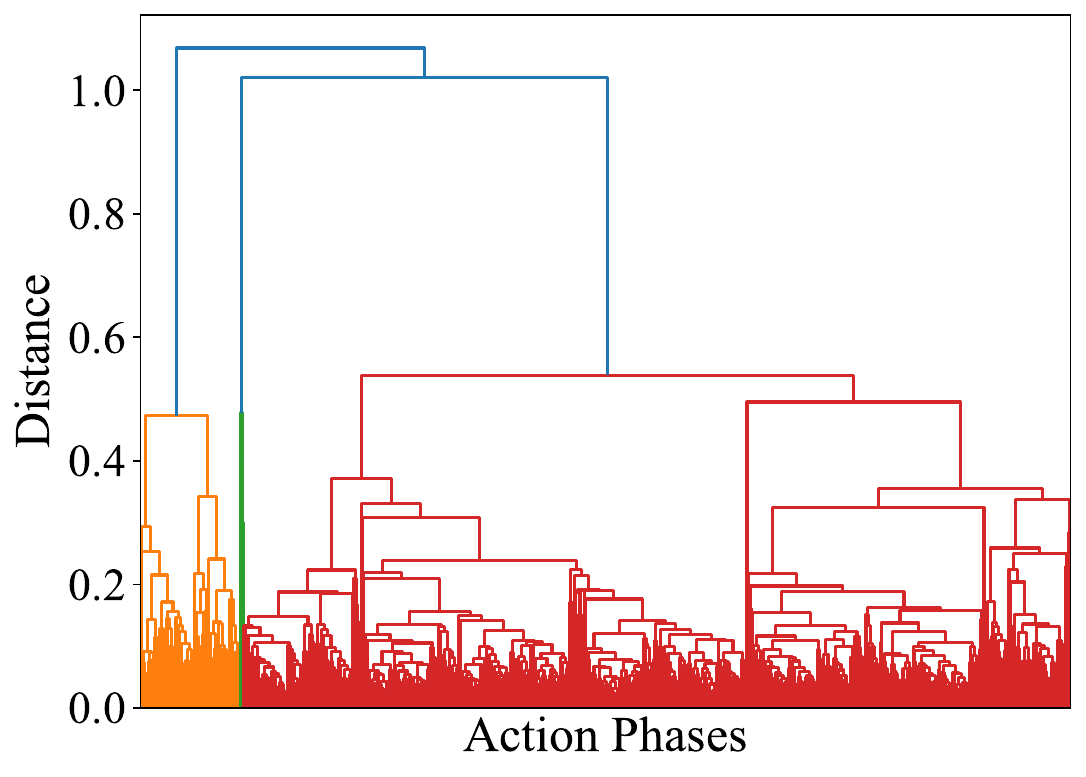}
    \caption{I80 dataset.}\label{fig:2nd_Cluster-I80}
  \end{subfigure}
  \hspace{0.2cm}
  \begin{subfigure}{0.47\textwidth}
  \centering
    \includegraphics[width=\textwidth]{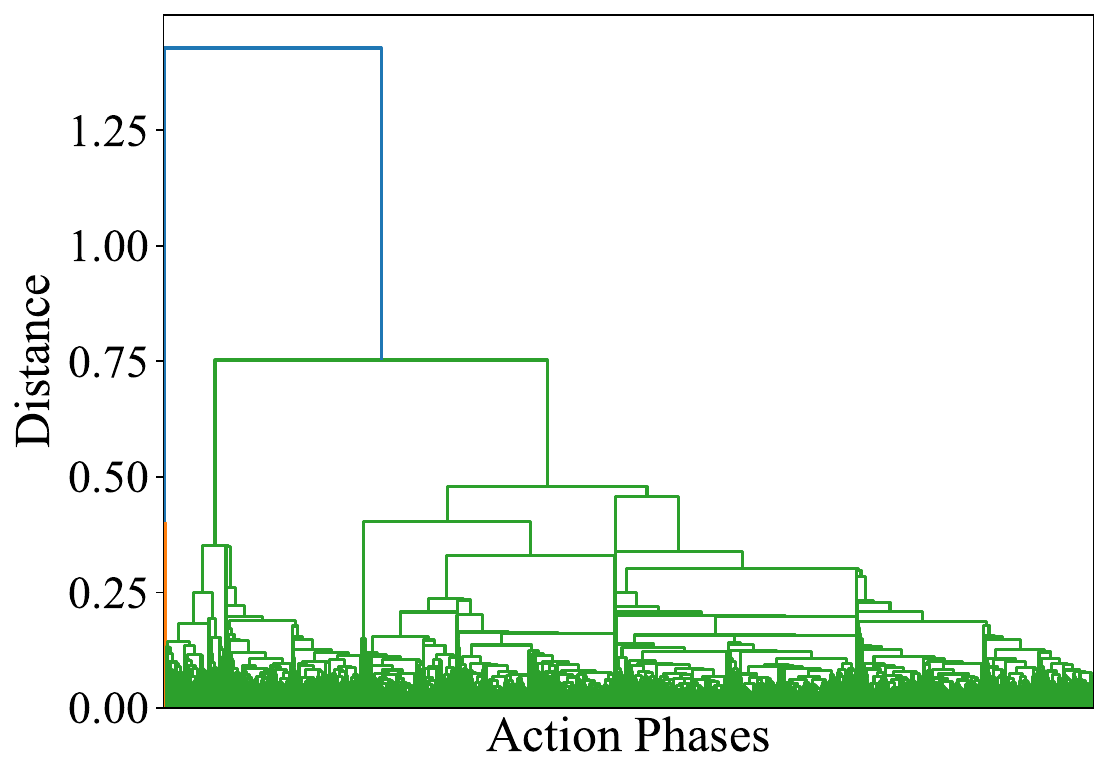}
    \caption{US101 dataset.}\label{fig:2nd_Cluster-US101}
  \end{subfigure}
  \caption{Dendrogram representations of hierarchical clustering results.}\label{fig:2ndClustering}
\end{figure}

Different from the initial round of analysis where feature selection is based on an unsupervised learning method, this round of analysis employs features selected based on the variable importance score (SI) computed in the prior round. Hierarchical clustering is also executed, with results depicted in Figure~\ref{fig:2ndClustering}. An observation of the dendrograms reveals three clusters in the I80 dataset and two clusters in the US101 dataset (see the sub-trees with different colors). Notice that sub-trees in each cluster are with small distances, or df, significantly smaller than the threshold $\delta = 1$, indicating high similarity among \emph{Action phases} within the cluster. Therefore, the results are directly considered as patterns obtained in this round of analysis. Combined with the updated cluster from the previous iterative analysis, the total number of clusters is obtained for each dataset, with each cluster representing a unique \emph{driving pattern}. 

Finally, 6 clusters in I80 dataset and 4 clusters in US101 dataset are found through the clustering calibration process. As the finding mentioned in \citep{YaoAction2023}, the traffic flow in US101 dataset exhibits less heterogeneity than that in the I80 dataset. Likewise, a smaller number of \emph{driving patterns} are identified in the latter dataset compared to the former. This consistency provides a connection between \emph{driving patterns} and the description of driving heterogeneity.

%% file: sections/4_DrivingpatternInterpretation.tex
\emph{Driving patterns} are represented by the final results of the clustering calibration process, and the interpretation for each pattern is provided in this section. Based on a measurement to distinguish different \emph{driving patterns}, i.e., the slope, a general analysis of the results is first presented. Then \emph{driving patterns} obtained from I80 dataset and US101 dataset are analyzed, respectively. 

\subsection{General Analysis of Driving Patterns}
The concept of ``Trend'' has consistently been a key of our \emph{Action phases} analysis, and it forms the basis for characterizing \emph{driving patterns}. Trends are illustrated through the slope, symbolizing the rate of change in a given variable. A positive slope portrays an increasing trend, such as an increase in velocity, while a negative slope implies a decreasing trend. The magnitude of the slope reflects the rate of change. Especially, several adjacent gentle slopes form fluctuations, representing a `Keeping' trend of variables. Given that variables usually manifest identical trends in different ways, such as linear increase, convex/concave progression, or slightly fluctuating increase, linear regression may struggle to precisely identify variable trends. To capture local trends within specified intervals in the dataset and retain overall trend accuracy, we employ a `sliding window' method. This technique involves forming a `window' of a particular size (e.g., 5 periods, 10 periods, etc.) which `slides' over the data points in the series \cite{chu1995time}. A linear regression is computed at each window position and the slope of the regression line is recorded. The final slope of the variable data, which serves as the trend index for each variable, is derived by averaging the slopes of these windows.

Figure~\ref{fig:variable-meanings-I80} and Figure~\ref{fig:variable-meanings-US101} use boxplots to illustrate the statistical data of the variable trend index for both datasets, respectively. Each boxplot displays the upper quartile, lower quartile, and median of the trend indexes. The whiskers extend to the farthest data points not deemed outliers, while outliers (if present) are denoted by asterisks. As previously mentioned, velocity holds the highest importance score when it comes to identifying dissimilarities within \emph{Action phases}, thus it is prioritized during analyzing the characteristics of \emph{driving patterns}. Given that the speed difference $\Delta v$ signifies the interaction between vehicles and their preceding vehicles, it is also taken into account when describing \emph{driving patterns}. Based on domain knowledge, if the velocity increases while the speed difference decreases, this represents the situation where the target vehicle is catching up with the vehicle in front. Conversely, a decrease in velocity accompanied by an increase in speed difference represents a situation of keeping away from the preceding vehicle. If both the velocity and speed difference remain roughly constant, it indicates that the vehicle is maintaining its current state.

\subsection{Driving Patterns in I80 dataset}
\begin{figure}[!ht]
  \centering
  \includegraphics[width=0.98\textwidth]{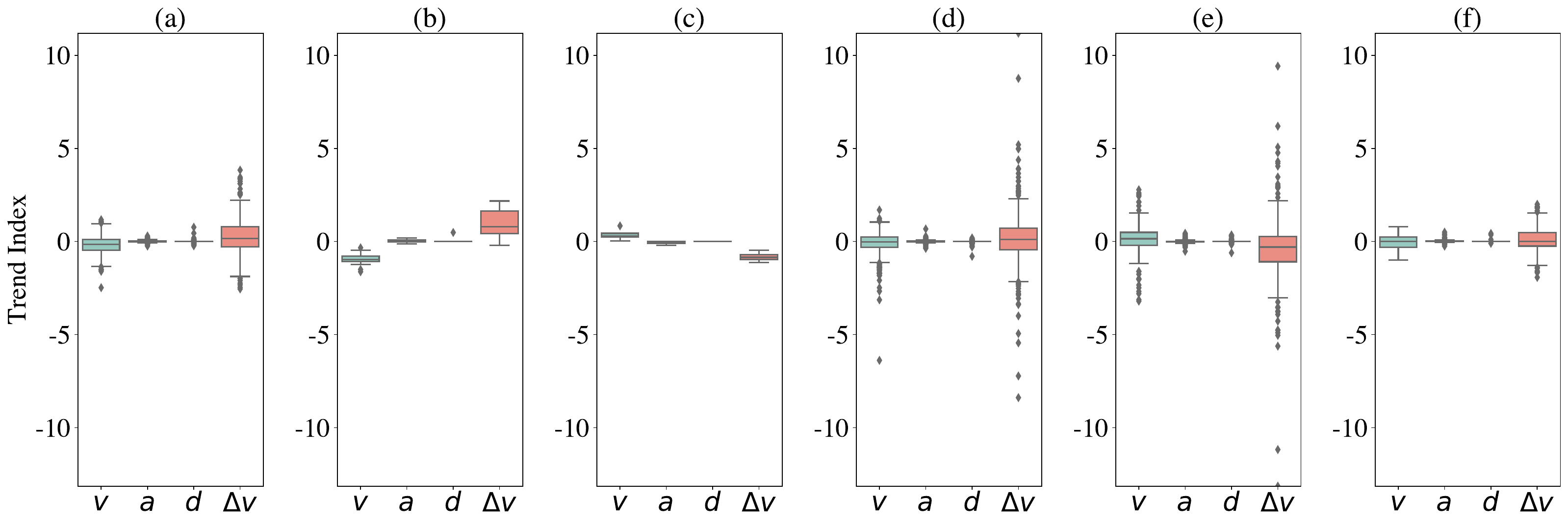}
  \caption{Trend index of variables in each cluster - I80 dataset.}\label{fig:variable-meanings-I80}
\end{figure}

Figure~\ref{fig:variable-meanings-I80} displays the trend index of variables for each cluster using the I80 dataset, where each sub-figure corresponds to a \emph{driving pattern}. Specifically, Figures~\ref{fig:variable-meanings-I80}(a)-(c) represent patterns derived from the first round of clustering, while Figures~\ref{fig:variable-meanings-I80}(d)-(f) represent those from the second round. Notably, in the cluster illustrated by Figure~\ref{fig:variable-meanings-I80}(c), the trend index of velocity for all \emph{Action phases} exceeds 0, indicating an increasing trend. Concurrently, the speed difference demonstrates a pronounced downward trend. Given the aforementioned domain knowledge, we label \emph{Action phases} in this cluster as the ``Catch up'' pattern. Conversely, the cluster of \emph{Action phases} depicted in Figure~\ref{fig:variable-meanings-I80}(b) exhibits a completely reverse trend, that is, a negative velocity trend index and a positive speed difference trend index, representing a \emph{driving pattern} named ``Keep away''. In Figure~\ref{fig:variable-meanings-I80}(f), the trend indexes of velocity and speed difference generally fluctuate around 0, denoting a ``Maintain distance'' pattern.

In the same way, the three aforementioned \emph{driving patterns} can also be identified in the other three figures. More specifically, Figure~\ref{fig:variable-meanings-I80}(e) illustrates an uptrend in velocity and a downtrend in speed difference, indicating a ``Catch up'' pattern. The ``Keep away'' and ``Maintain distance'' patterns can be discerned in Figures~\ref{fig:variable-meanings-I80}(a) and~\ref{fig:variable-meanings-I80}(d), respectively. It is worth noting that the patterns manifest with more instability in these three clusters, thus being labeled as an Unstable state. On the contrary, the \emph{driving patterns} detected in Figures~\ref{fig:variable-meanings-I80}(c),~\ref{fig:variable-meanings-I80}(b), and~\ref{fig:variable-meanings-I80}(f) reflect a Stable state. Ultimately, the \emph{driving patterns} identified in the I80 dataset are interpreted as ``Stable catch up'', ``Stable keep away'', ``Stable maintain distance'', ``Unstable catch up'', ``Unstable keep away'', and ``Unstable maintain distance'', as summarized in Table~\ref{tab:patterns-80}. The shade of color corresponds to the size of each pattern. Generally, Unstable patterns considerably outweigh Stable ones in size. Among the Stable patterns, ``Maintain distance'' exceeds the other two dynamic patterns in size. This observation aligns with empirical knowledge: driving is a dynamic process and contains stochastic, maintaining the status is the simplest approach to perform a Stable state. 

\begin{table}[!ht]
	\caption{Overview of \emph{driving patterns} - I80 dataset}\label{tab:patterns-80}
	\begin{center}
		\begin{tabular}{p{2cm}  p{3cm}  p{3cm}  p{3cm}}
		 & Catch up & Keep away & Maintain distance \\\hline
		Stable & \cellcolor{red!4} (c) & \cellcolor{red!8} (b) & \cellcolor{red!20} (f)  \\
	Unstable & \cellcolor{red!40} (e)  & \cellcolor{red!60} (a) & \cellcolor{red!52} (d) \\\hline
		\end{tabular}
	\end{center}
\end{table}

\subsection{Driving Patterns in US101 dataset}
\begin{figure}[!ht]
  \centering
  \includegraphics[width=0.75\textwidth]{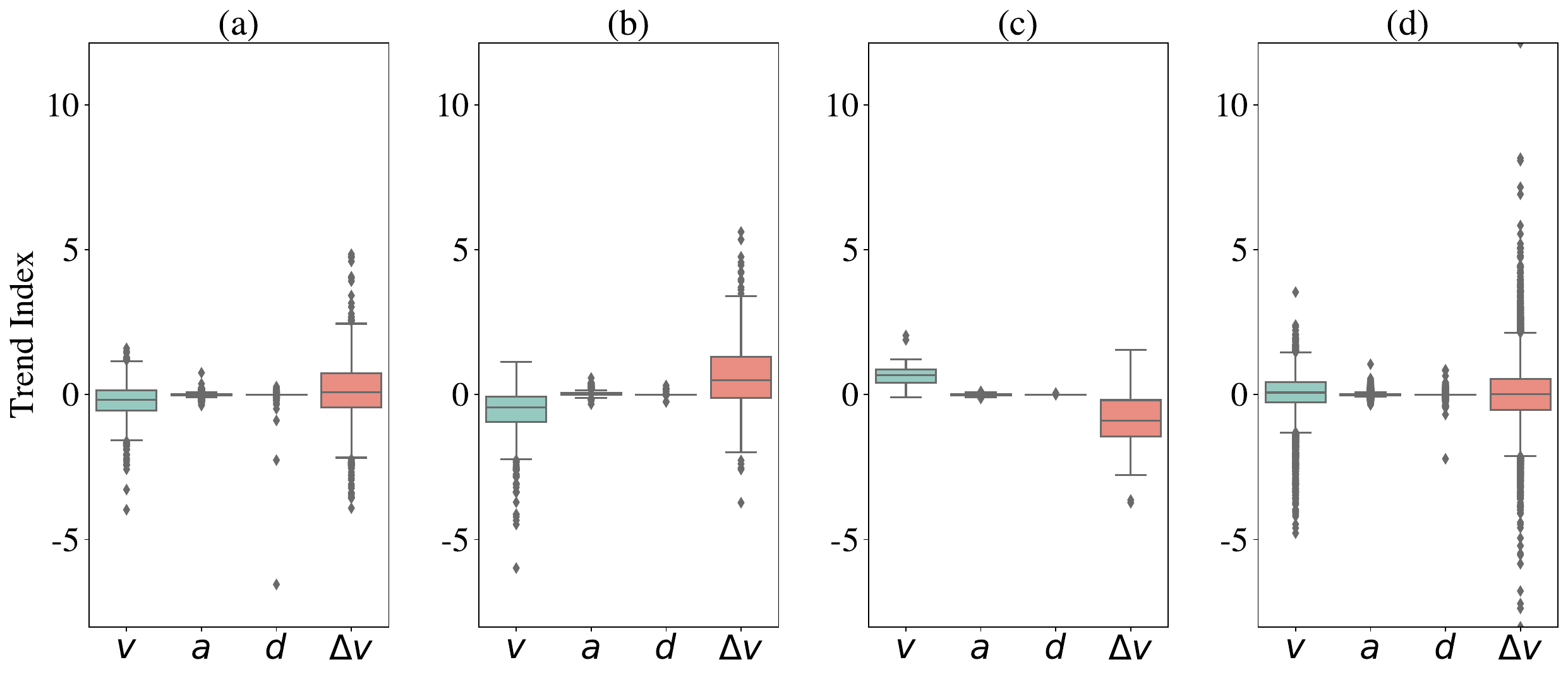}
  \caption{Trend index of variables in each cluster - US101 dataset.}\label{fig:variable-meanings-US101}
\end{figure}

Corresponding \emph{driving patterns} are also identified in the US101 dataset as illustrated in Figure~\ref{fig:variable-meanings-US101}. In Figure~\ref{fig:variable-meanings-US101}(c), most of the \emph{Action phases} display an increasing trend in velocities. Conversely, the overall speed difference exhibits a decreasing trend. Considering the limited outliers in both variables' indexes, this pattern is interpreted as a ``Stable catch up'' pattern. Figure~\ref{fig:variable-meanings-US101}(b) demonstrates an obvious downward trend in velocity and an upward trend in speed difference, coupled with numerous outliers. Hence, this pattern is recognized as ``Unstable keep away''. Both Figure~\ref{fig:variable-meanings-US101}(a) and Figure~\ref{fig:variable-meanings-US101}(d) exhibit a ``Maintain distance'' pattern. As the outliers in Figure~\ref{fig:variable-meanings-US101}(d) are significantly more than those in Figure~\ref{fig:variable-meanings-US101}(a), they are labeled as ``Unstable'' and ``Stable'', respectively. As the statistics presented in Table~\ref{tab:patterns-101} reveal, the ``Maintain distance'' pattern has the largest size, where the Unstable pattern significantly outweighs the Stable one. The \emph{driving pattern} with the smallest size in this dataset is ``Stable catch up''. Notice that patterns of ``Stable keep away'' and ``Unstable catch up'' observed in the I80 dataset are absent here, making the ``Maintain distance'' pattern significantly surpasses others in frequency. This is consistent with driving behavior in the relatively heavy traffic observed during morning peak hours.

\begin{table}[!ht]
	\caption{Overview of \emph{driving patterns} - US101 dataset}\label{tab:patterns-101}
	\begin{center}
		\begin{tabular}{ p{2cm}  p{3cm}  p{3cm}  p{3cm}}
		 & Catch up & Keep away & Maintain distance \\\hline
		Stable & \cellcolor{red!3} (c) &   & \cellcolor{red!10} (a)  \\
	Unstable &   & \cellcolor{red!25} (b) & \cellcolor{red!60} (d) \\\hline
		\end{tabular}
	\end{center}
\end{table}

%% file: sections/5_Conclusion.tex
To capture representations of driving characteristics and facilitate a comprehensive understanding of driving behavior, this study proposed a framework to cluster \emph{Action phases} and interpret those clusters as several \emph{driving patterns}. This section presents a summary of the key findings and main conclusions derived from this study. Then discussion and outlook of the proposed framework are provided to shed light on its implications and future directions.

\subsection{Findings of the Study}
As clustering algorithms need input data arrays of equal length, a Resampling and Downsampling Method (RDM) was first adopted to standardize the various length of \emph{Action phases} in this framework. Then, the clustering calibration process of ``Feature Selection'', `Clustering Analysis'', ``Difference/Similarity Evaluation'', and ``\emph{Action phases} Re-extraction'' was iterated until all differences among clusters and similarities within clusters meet the pre-determined criteria. Finally, six clusters were observed in I80 dataset, indicating six \emph{driving patterns}, which have been labeled as ``Catch up'', ``Keep away'', and ``Maintain distance'', each with ``Stable'' and ``Unstable'' states. These \emph{driving patterns} were also identified in the US101 dataset, while the patterns ``Stable keep away'' and ``Unstable catch up'' were absent. 

\subsection{Conclusions}
The main conclusions of this study are summarized below:

(i) Velocity $v$ exhibits the highest importance score among the considered four variables, suggesting that it reflects more characteristics of driving behavior.

(ii) In general, Unstable patterns significantly outnumber Stable ones in terms of size. Among the Stable patterns, ``Maintain distance'' exceeds the other two dynamic patterns in size. This observation aligns with empirical knowledge: driving is a dynamic and stochastic process, and maintaining the status is the simplest way to achieve a Stable state.

(iii) Comparable \emph{driving patterns} have been detected in both the I80 and US101 datasets. Notably, the patterns ``Stable keep away'' and ``Unstable catch up'' are missing from the US101 dataset. As previously identified by \citep{YaoAction2023}, the traffic flow in US101 dataset displays less heterogeneity compared to the I80 dataset. This consistency indicates the prospective advantages of using \emph{driving pattern} to illustrate the heterogeneity in driving behavior.

\subsection{Discussion and Outlook}
This framework introduces an unsupervised learning method to improve the manual categorization of heterogeneous driving behaviors, thereby addressing the pitfalls of depending solely on experiential knowledge. By incorporating a wide array of driving behavior characteristics, such as stability and driving state, more accurate and justifiable labels result. This will help to alleviate the scarcity of labels in supervised learning and consequently bolster its performance in tasks such as driving behavior modeling and driving trajectory prediction, among others. However, the framework does have its limitations, which are the focus of our future research. First, the framework needs further validation to strengthen the credibility of the derived \emph{driving patterns}. Additionally, the methods and techniques employed in each element of the framework could be further optimized and justified.